\documentclass{article}

\usepackage[final]{neurips_2025}

\usepackage[utf8]{inputenc} 
\usepackage[T1]{fontenc}    
\usepackage{hyperref}       
\usepackage{url}            
\usepackage{booktabs}       
\usepackage{amsfonts}       
\usepackage{nicefrac}       
\usepackage{microtype}      
\usepackage{xcolor}         
\usepackage{graphicx}
\usepackage{amsmath} 
\usepackage{cleveref} 
\newtheorem{theorem}{Theorem}
\newtheorem{remark}{Remark}
\newtheorem{proposition}{Proposition}

\newcommand{\bs}{\boldsymbol} 
\newcommand{\PP}{\mathbb{P}}

\usepackage{algorithm2e} 
\crefname{algocf}{algorithm}{algorithms} 
\usepackage{csquotes} 
\usepackage{dsfont}
\usepackage{listings}
\usepackage{xcolor}
\usepackage{subcaption}
\usepackage[export]{adjustbox}

\title{Scalable and adaptive prediction bands with kernel sum-of-squares}

\author{%
Louis Allain$^{1,2}$ \quad S\'ebastien Da Veiga$^{2}$ \quad Brian Staber$^1$ \\
$^1$Safran Tech, Digital Sciences \& Technologies,78114 Magny-Les-Hameaux, France \\ $^2$Univ Rennes, Ensai, CNRS, CREST - UMR 9194, F-35000 Rennes, France\\
\texttt{\{louis.allain,brian.staber\}@safrangroup.com}\\
\texttt{sebastien.da-veiga@ensai.fr}
}

\begin{document}

\maketitle

\begin{abstract}
Conformal Prediction (CP) is a popular framework for constructing prediction bands with valid coverage in finite samples, while being free of any distributional assumption. A well-known limitation of conformal prediction is the lack of adaptivity, although several works introduced practically efficient alternate procedures. In this work, we build upon recent ideas that rely on recasting the CP problem as a statistical learning problem, directly targeting coverage and adaptivity. This statistical learning problem is based on reproducible kernel Hilbert spaces (RKHS) and kernel sum-of-squares (SoS) methods. First, we extend previous results with a general representer theorem and exhibit the dual formulation of the learning problem. Crucially, such dual formulation can be solved efficiently by accelerated gradient methods with several hundreds or thousands of samples, unlike previous strategies based on off-the-shelf semidefinite programming algorithms. Second, we introduce a new hyperparameter tuning strategy tailored specifically to target adaptivity through bounds on test-conditional coverage. This strategy, based on the Hilbert-Schmidt Independence Criterion (HSIC), is introduced here to tune kernel lengthscales in our framework, but has broader applicability since it could be used in any CP algorithm where the score function is learned. Finally, extensive experiments are conducted to show how our method compares to related work. All figures can be reproduced with the accompanying code at \href{https://www.gitlab.com/drti/ksos-bands}{gitlab.com/drti/ksos-bands}.
\end{abstract}

\section{Introduction}
\label{sec:introduction}
In many applications, machine learning regression models require a trustworthy uncertainty quantification in their predictions. This is especially true for high-stakes applications such as design optimization, non-destructive testing, medical diagnostics, autonomous vehicles, or financial forecasting, where decisions based on model predictions can have significant impacts.
Having a reliable uncertainty quantification is thus fundamental. Several machine learning models come with uncertainty quantification in their predictions, such as Gaussian Processes \citep{rasmussen2005gaussianprocessesmachinelearning}, Random Forests \citep{breiman2001randomforests} or Bayesian Neural Networks \citep{wang2020surveybayesiandeeplearning}, among many others, but these models generally provide inaccurate prediction bands: coverage guarantees typically hold asymptotically or with strong distributional assumptions. In practice, however, especially for high-stakes decisions, we should at least provide marginal coverage guarantees that hold in finite sample and without making any distributional assumptions on the data. In addition, a desirable feature is adaptivity: we would like prediction bands to be wide when either the model lacks confidence or if the variability in the data is high, and narrow when both the model is confident and the variability is low. Having adaptive prediction bands means having an uncertainty quantification that is informative on either the performance of the model or the variability of the data, which is key in crucial applications.

Conformal Prediction (CP) (see, e.g., \citep{gammerman1998learningbytransduction, papadopoulos2002inductiveconfidencemachineregression, shafer2007tutorialconformalprediction} or \citep{angelopoulos2022gentleintroductionconformalprediction} for a modern introduction) has been designed from the ground up to be an uncertainty quantification statistical framework that provides marginal coverage guarantees in finite sample while being distribution-free. In particular, the split conformal procedure proposed by \citet{papadopoulos2002inductiveconfidencemachineregression} is especially easy to implement. CP is becoming widely used in many different applications (see \citep{balasubramanian2014conformalpredictionforreliablemachinelearningtheoryadaptationsapplications, vazquez2022conformalpredictionclinicalmedicalsciences} and references therein), but unfortunately, by construction, standard CP does not provide adaptive prediction bands. A lot of research has been done in this direction, which we will review later.

In parallel, for specific statistical learning problems, \citet{marteauferey2020nonparametricmodelsnonnegativefunctions} introduced a new kernel framework known as kernel sum-of-squares (SoS), tailored specifically to estimate non-negative functions. Their key idea is to characterize such functions of interest by a linear positive semidefinite Hermitian operator, which admits a finite-dimensional representation through a representer theorem. Since then, it has been leveraged for non-convex optimization \citep{rudi2020findingglobalminimakernel}, estimation of optimal transport distances \citep{vacher2021dimensionfreecomputationalupperboundsmooth}, modeling of probability densities \citep{rudi2021psdrepresentationseffectiveprobability} and PSD-constrained functions \citep{muzellec2022learningpsdvaluedfunctionsusing}. Very recently, kernel SoS was also identified as a powerful framework for constructing more adaptive prediction bands by  \citet{liang2022universalpredictionbandssemidefiniteprogramming} and \citet{fan2024utopiauniversallytrainableoptimal}. They were the first to propose to build prediction bands as solutions of such a learning problem, where adaptive coverage is targeted with additional constraints. This point of view is the one we adopt and generalize in this paper.

\paragraph{Outline and contributions.}
We start by presenting CP in \Cref{sec:conformal prediction_kSos} as well as recent variants developed to improve adaptivity. We also introduce the kernel SoS framework, since our proposed method extensively relies on it. In \Cref{sec:our method} we introduce our approach that learns a CP score function by solving a statistical optimization problem with several ingredients: an objective function controlling both the width and the regularity of the prediction bands, and constraints for coverage. We also discuss a new criterion dedicated to the tuning of kernel lengthscales, with local coverage as an objective. In \Cref{sec:experiments}, we finally conduct extensive experiments to compare our method to other conformal prediction methods that provide adaptive bands.

Our contributions are as follows:
\begin{itemize}
    \item We generalize the previously introduced kernel SoS point of view for prediction bands and precisely analyze the contribution and practical effect of each term in the objective function,
    \item We provide a representer theorem that makes the problem numerically tractable,
    \item We derive a dual formulation of this problem and propose an accelerated gradient algorithm to enable faster computation on large datasets, unlike previous work that was limited to small datasets,
    \item We introduce a new criterion to tune kernel hyperparameters based on the Hilbert-Schmidt Independence Criterion (HSIC), which is also applicable to any other CP method. In particular, we provide both theoretical and empirical evidence of the effectiveness of this metric to achieve better adaptivity.
\end{itemize}

\section{Conformal prediction and kernel sum-of-squares}
\label{sec:conformal prediction_kSos}
\subsection{Conformal prediction}

\paragraph{Split conformal prediction.} CP was introduced by \citet{gammerman1998learningbytransduction}, with the so-called full-variant, but we focus here on the split variant, introduced by \citet{papadopoulos2002inductiveconfidencemachineregression}. Suppose we have a training dataset \(\mathcal{D}_N = \{\left(X_i, Y_i\right)\}_{i=1}^{N}\) from a pair \(\left(X, Y\right)\sim P_{XY}\) where \(X\in\mathcal{X}\subset\mathbb{R}^d\) and \(Y\in\mathcal{Y}\subset\mathbb{R}\). This dataset is split in two parts: a \emph{pre-training} dataset \(\mathcal{D}_n = \{\left(X_i, Y_i\right)\}_{i=1}^{n}\) and a \emph{calibration} one \(\mathcal{D}_m = \{\left(X_i, Y_i\right)\}_{i=1}^{m}\) with $N=n+m$.

The pre-training dataset \(\mathcal{D}_n\) is used to first fit a predictive model \(\widehat{m}_n(\cdot)\), which can be any machine learning algorithm. The second step consists in computing performance scores associated to \(\widehat{m}_n\) on the hold-out calibration dataset \(\mathcal{D}_m\). The usual score in the literature is defined as the absolute errors \(S(X_i,Y_i):=S_i = \lvert Y_i-\widehat{m}_n(X_i) \rvert\) for \(i \in \mathcal{D}_m\), which are used to compute the quantile \(\widehat{q}_{\alpha}\) of the set \(\{S_i\}_{i \in \mathcal{D}_m}\) with an adjusted level \(\lceil(1-\alpha)(m+1)\rceil/m\), where \(\alpha\) is the desired error rate. Finally, for a new observation \(X_{N+1}\), the split CP prediction bands are \(\widehat{C}_{N}(X_{N+1}) = \left[\widehat{m}_n(X_{N+1}) \pm \widehat{q}_{\alpha}\right]\), which satisfy the marginal coverage
\begin{align}
    \label{eq:marginal_coverage}
    \PP\left(Y_{N+1} \in \widehat{C}_{N}(X_{N+1})\right) \geq 1 - \alpha
\end{align}
for any $N$ as long as \(\left(X_1, Y_1\right),\ldots,\left(X_N, Y_N\right),\left(X_{N+1}, Y_{N+1}\right)\) are exchangeable.
Unfortunately, as underlined by \citet{romano2019conformalizedquantileregression}, these prediction bands cannot be adaptive since they have constant width \(2\widehat{q}_{\alpha}\). The research direction in recent years has been to adjust CP procedures to target more adaptive bands.

\paragraph{The quest for adaptivity.}
Historically the first idea was to change the score function by rescaling the scores with an estimate of the variability \(\widehat{\sigma}_n(\cdot) \geq 0\). The new scores are thus defined as  \(
    S_i = \lvert Y_i-\widehat{m}_n(X_i)\rvert/\widehat{\sigma}_n(X_i) , \; i\in \mathcal{D}_m \) with prediction bands \(\widehat{C}_{N}(X_{N+1}) = \left[\widehat{m}_n(X_{N+1}) \pm \widehat{q}_{\alpha}\widehat{\sigma}_n(X_{N+1})\right]\).
\citet{lei2012distributionfreepredictionbands} first proposed to use an estimate of the conditional mean absolute deviation for \(\widehat{\sigma}_n(\cdot)\). Another sensible choice is to scale the scores by an estimate of the standard deviation, as was suggested for several machine learning models like Gaussian Processes, Random Forests and Bayesian Neural Networks \citep{johansson2014regressionconformalpredictionrandomforests, papadopoulos2023guaranteedcoveragepredictionintervalsgaussianprocessregression, jaber2024conformalapproachgaussianprocess}.
However, such scaling functions are rarely estimated in a goal-oriented way, without quantitative and explicit consideration for adaptive coverage. This often leads in practice to poorly adaptive prediction bands.
In a parallel line of work, the popular Conformalized Quantile Regression (CQR) \citep{romano2019conformalizedquantileregression} proposes to change the score function by leveraging quantile regression. Instead of using an interval built around an estimate \(\widehat{m}_n(\cdot)\) of the regression function, they rely on estimates \(\widehat{q}^{\alpha_{lower}}_n(\cdot)\) and \(\widehat{q}^{\alpha_{upper}}_n(\cdot)\) of the conditional quantiles, and build an interval \(
\widehat{C}_{N}(X_{N+1}) = \left[q^{\alpha_{lower}}_n(X_{N+1}) - \widehat{q}_{\alpha},q^{\alpha_{upper}}_n(X_{N+1}) + \widehat{q}_{\alpha}\right] \)
where now \(\widehat{q}_{\alpha}\) is the adjusted quantile of the set \(\{\max\left(q^{\alpha_{lower}}_n(X_i)-Y_i,Y_i-q^{\alpha_{upper}}_n(X_i) \right), \; i\in \mathcal{D}_m\}\). In other words, the score function is chosen as \(S(X,Y)=\max\left(q^{\alpha_{lower}}_n(X)-Y,Y-q^{\alpha_{upper}}_n(X) \right)\). By design, the CQR model is adaptive and generally provides sensible prediction bands. However, it suffers from two practical limitations: (a) decision-making people usually prefer a point estimate with an interval \emph{around} this estimate and (b) quantile regression in a small data regime can be quite challenging. Also note that a regularized version of CQR tailored to target test-conditional coverage was recently proposed by \citet{feldman2021improvingconditionalcoverageorthogonal}.

\smallskip

Another line of work consists in modifying the calibration step. For example, \citet{guan2022localizedconformalpredictiongeneralized} and \citet{ hore2024conformalpredictionlocalweights} propose to weight the scores when computing the adjusted level quantile, where the weights depend on the test point \(X_{N+1}\): this directly implies that the quantile changes with \(X_{N+1}\), and as a result the prediction bands are adaptive. More precisely, given a kernel \(H(\cdot, \cdot)\) that defines a density \(H(x, \cdot)\) for all \(x\in \mathcal{X}\), sample \(\widetilde{X}_{N+1}\) from \(H(X_{N+1}, \cdot)\). The quantile \(\widehat{q}_{\alpha}(X_{N+1}, \widetilde{X}_{N+1})\) is computed on the empirical distribution \(\sum_{i=1}^{m}\widetilde{w}_{i}\delta_{S_i}+\widetilde{w}_{N+1}\delta_{+\infty}\), where the weights are computed as \(\widetilde{w}_{i} = H(X_i,\widetilde{X}_{N+1})/(\sum_{j=1}^{m}H(X_j, \widetilde{X}_{N+1})+H(X_{N+1}, \widetilde{X}_{N+1}))\). Although appealing, such modifications to the score function have some practical shortcomings. First, computing different weights for all test points can be computationally demanding during inference. Second, the method suffers in practice from the randomization induced by the sampling of \(\widetilde{X}_{N+1}\). To overcome this issue \citet{hore2024conformalpredictionlocalweights} propose the \(m\)-RLCP methods that averages the predictions bands over \(m\) sampling of \(\widetilde{X}_{N+1}\). However this leads to a marginal coverage equal to \(1-2\alpha\) and the computational cost increases significantly. Finally, and perhaps more importantly, the kernel \(H(\cdot, \cdot)\) involved in the definition of the weights depends on a bandwidth hyperparameter that must be tuned. This choice has a strong  impact on the shape of the prediction bands as shown in \citet{hore2024conformalpredictionlocalweights}. We will come back to this point later since our framework shares the same characteristic.
Finally, another reweighting method based instead on Jackknife+ was proposed by \citet{deutschmann2023adaptiveconformalregressionjackknife}.

\subsection{Kernel sum-of-squares}
\label{sec:kernel sum-of-squares}
Since our proposed method is based on kernel SoS for positive functions, we give a brief overview following \citet{marteauferey2020nonparametricmodelsnonnegativefunctions}.

Let \(\mathcal{H}\) be a RKHS with associated kernel \(k\) and \(\phi\colon \mathcal{X} \rightarrow \mathcal{H}\) one of its feature map such that \(k(x,x')=\phi(X)^{\top}\phi(X)\). Let \(\mathcal{S}(\mathcal{H})\) be the set of bounded Hermitian linear operator from \(\mathcal{H}\) to \(\mathcal{H}\). For \(\mathcal{A} \in \mathcal{S}(\mathcal{H})\), we write \(\mathcal{A} \succeq 0\) when \(\mathcal{A}\) is a positive semi-definite (PSD) operator, and \(\mathcal{S}_{+}(\mathcal{H})\) the set of such PSD operators. For all \(x\in \mathcal{X}\), we define \(f_{\mathcal{A}}(X) = \phi(X)^{\top}\mathcal{A}\phi(X)\), \(\mathcal{A}\in \mathcal{S}_{+}(\mathcal{H})\) which is non-negative by construction. Kernel SoS refers to a statistical learning problem where the unknown nonparametric function is constrained to be non-negative and obtained as the solution of
\begin{align}
\label{equation:general problem marteau-ferey}
    \inf_{\mathcal{A}\succeq 0} L(f_{\mathcal{A}}(X_1), \ldots, f_{\mathcal{A}}(X_n)) + \lambda_{1}\lVert \mathcal{A}\rVert_{\star} + \lambda_{2}\lVert\mathcal{A}\rVert_{\text{F}}^{2}
\end{align}
where \(\lVert\mathcal{A}\rVert_{\star}\) and \(\lVert\mathcal{A}\rVert_{F}\) denote the nuclear norm and the Frobenius norm of operator \(\mathcal{A}\), respectively. Interestingly, \citet{marteauferey2020nonparametricmodelsnonnegativefunctions} show a representer theorem for \Cref{equation:general problem marteau-ferey}, which makes kernel SoS computationally tractable in finite-dimension and is recalled below.
\begin{theorem}[\citet{marteauferey2020nonparametricmodelsnonnegativefunctions}]
\label{thm:representer theorem marteau ferey}
    Assume \(L\colon \mathbb{R}^{n}\rightarrow \mathbb{R}\cup\{+\infty\}\) to be a lower semi-continuous and bounded below loss function. \Cref{equation:general problem marteau-ferey} admits a solution \(\mathcal{A}^{\star}\) which can be written \(\mathcal{A}^{\star} = \sum_{i,i=1}^{n}B^{\star}_{ij}\phi(X_i)\phi(X_j)^{\top}\) for some matrix \(\mathbf{B}^{\star}\in \mathbb{R}^{n\times n},\mathbf{B}^{\star}\succeq0\). Furthermore \(\mathcal{A}^{\star}\) is unique if \(L\) is convex and \(\lambda_2 >0\). The corresponding non-negative function is given by \(f_{ \mathcal{A}^{\star}}(X) = \sum_{i,i=1}^{n}B^{\star}_{ij}k(X_i, X)k(X_j, X)\).
\end{theorem}

\Cref{thm:representer theorem marteau ferey} provides a finite-dimensional equivalent problem which involves an unknown PSD matrix \(\mathbf{B}\). But  \citet{marteauferey2020nonparametricmodelsnonnegativefunctions} also propose an equivalent formulation: considering \(\mathbf{K}\) the kernel matrix with elements \(K_{ij}=k(X_i,X_j)\) and  \(\mathbf{V}\) its upper Cholesky decomposition, we can define \( \bs{\Phi}(X) = \mathbf{V}^{-\top}\mathbf{k}_X\) with \(\mathbf{k}_X=(k(X,X_i))_{i=1,\ldots,n}\) and \(\tilde{f}_{\mathbf{A}}(X) = \bs{\Phi}(X)^{\top}\mathbf{A} \bs{\Phi}(X)\). With these notations, the following proposition shows that we obtain the same solution if we optimize the PSD matrix \(\mathbf{A}\) instead of \(\mathbf{B}\).
\begin{proposition}[\citet{marteauferey2020nonparametricmodelsnonnegativefunctions}]
     Under the assumptions of \Cref{thm:representer theorem marteau ferey}, the following problem has at least one solution, which is unique if \(\lambda_2 >0\) and \(L\) is convex:
    \begin{align}
    \label{problem:equivalent representer theorem}
        \inf_{\mathbf{A}\succeq 0} L(\tilde{f}_{\mathbf{A}}(X_1), \ldots, \tilde{f}_{\mathbf{A}}(X_n)) + \lambda_{1}\lVert \mathbf{A}\rVert_{\star} + \lambda_{2}\lVert\mathbf{A}\rVert_{\text{F}}^{2}.
    \end{align}
    For any given solution \(\mathbf{A}^{\star}\in\mathbb{R}^{n\times n}\) of \Cref{problem:equivalent representer theorem}, the function \(\tilde{f}_{\mathbf{A}^{\star}}\) is also  solution of \Cref{equation:general problem marteau-ferey}.
\end{proposition}
Although such a result may appear of minor impact, the \(\mathbf{A}\) formulation actually yields significant computational savings in practice, as we illustrate in our numerical experiments (see Appendix B.4).

\begin{remark}
\label{rmk:sum-of-squares nomenclature}
    Operator \(\mathcal{A}\) is PSD, hence it admits an eigendecomposition \(\mathcal{A} = \sum_{l\geq0} \lambda_l u_l\otimes u_l\) with \(\lambda_l\geq 0\) and \(u_l\in\mathcal{H}\).
    By the reproducing property, we have \(f_{\mathcal{A}}(X)=\sum_{l\geq0} \lambda_l u_l(X)u_l(X)^{\top}\). Hence, \(f_{\mathcal{A}}(X)\) is an infinite sum of squared functions in \(\mathcal{H}\). Equivalently in finite dimension, for a PSD matrix \(\mathbf{B} = \mathbf{U}\mathbf{D}\mathbf{U}^{\top}\) we have \(f_{\mathbf{B}}(X)=\mathbf{k}_X^{\top}\mathbf{U}\mathbf{D}\mathbf{U}^{\top}\mathbf{k}_X = \sum_{l=1}^{n}\lambda_l (\sum_{i=1}^{n}u_{il}k(X, X_i))^{2}\). Thus \(f_{\mathbf{B}}\) is a function defined as a linear combination of squared functions from \(\mathcal{H}\).
\end{remark}

\section{Regularized kernel SoS for adaptive prediction bands}
\label{sec:our method}

We now come back to our initial problem of building adaptive prediction bands. To do so, we focus on the split CP setting with two i.i.d. datasets \(\mathcal{D}_n = \{\left(X_i, Y_i\right)\}_{i=1}^{n}\) and \(\mathcal{D}_m = \{\left(X_i, Y_i\right)\}_{i=1}^{m}\), and consider estimating a score function in a specific supervised learning problem to achieve better adaptivity.

\subsection{Learning the scores through an optimization problem}
A general framework for learning score functions was recently introduced by  \citet{xie2024boostedconformalpredictionintervals}:  given a task-specific loss (e.g. conditional coverage or minimum interval width), an optimized score function is obtained via a boosting algorithm. In this work, the score function \(S(X,Y)=\max\left(\mu_1(X)-Y,Y-\mu_2(X)\right)/\sigma(X)\) is inspired by CQR and is parameterized by three unknown functions \((\mu_1,\mu_2,\sigma)\) such that \(\mu_1(\cdot) \leq \mu_2(\cdot)\) and \(\sigma(\cdot) \geq 0\), which are iteratively optimized during boosting rounds. From a practical viewpoint however, the constraints on these functions are not inherently accounted for in the boosting algorithm.

We advocated before the use of prediction intervals built around a point estimate, which corresponds to the particular case \(m=\mu_1=\mu_2\) and yields a score function \(S(X,Y)= \vert Y-m(X)\vert/\sigma(X)\), or equivalently \(S(X,Y)= (Y-m(X))^2/f(X)\) with \(f(\cdot):=\sigma(\cdot)^2\) since only the score quantiles are involved in the final prediction interval: we thus recover the rescaled conformal score setting where we \textit{learn} the rescaling function \(f(\cdot)\), with an additional non-negativity constraint. The kernel SoS framework is thus a natural candidate for this learning task.

\paragraph{Kernel SoS formulation for prediction intervals.}
We introduce two RKHSs \(\mathcal{H}^m\) and \(\mathcal{H}^f\) associated to kernels \(k^m\) with lengthscales \(\theta^m\) and \(k^f\) with lengthscales \(\theta^f\), respectively. The regression function \(m\) will be estimated in the RKHS \(\mathcal{H}^m\) while the non-negative scaling function \(f\) will be estimated using the kernel SoS framework in the RKHS \(\mathcal{H}^f\). The first step is to derive our proposed infinite-dimensional learning problem from the properties that we impose on prediction bands:
\begin{align}
    \inf_{m\in \mathcal{H}^{m},\; \mathcal{A}\in \mathcal{S}_{+}\left(\mathcal{H}^{f}\right)} \quad&\frac{a}{n}\sum_{i=1}^{n} \left(Y_{i}-m(X_{i})\right)^{2} + \frac{b}{n}\sum_{i=1}^{n} f_{\mathcal{A}}(X_{i}) + \lambda_{1}\lVert\mathcal{A}\rVert_{\star} + \lambda_{2}\lVert\mathcal{A}\rVert_{F}^{2} \label{eq:infdim1} \\ 
    \text{s.t.} \quad& f_{\mathcal{A}}(X_{i}) \geq \left(Y_{i}-m(X_{i})\right)^{2}, \;i \in \left[n\right],\label{eq:infdim2} \\ 
    \quad& \lVert m\rVert_{\mathcal{H}^{m}}^{2} \leq s. \label{eq:infdim3}
\end{align}
In this problem, we propose to include several key components:
\begin{enumerate}
    \item Accurate mean estimation (first term in (\ref{eq:infdim1})) with regularity penalty (\ref{eq:infdim3}), as in standard kernel ridge regression \citep{scholkopf2002learningwithkernels},
    \item Prediction intervals with minimum mean width (second term in (\ref{eq:infdim1})) and an additional regularity penalty (third and fourth terms in (\ref{eq:infdim1})),
    \item \(100\%\) coverage on pre-training data (\ref{eq:infdim2}), later calibrated on the calibration dataset.
\end{enumerate}
Minimizing the nuclear norm with coverage constraints was originally proposed by \citet{liang2022universalpredictionbandssemidefiniteprogramming}, and minimizing the mean width was later proposed by \citet{fan2024utopiauniversallytrainableoptimal}. Our proposition essentially differs from their work for a broader and more efficient practical applicability: (a) we place ourselves in the split CP setting, whereas \citet{liang2022universalpredictionbandssemidefiniteprogramming} and \citet{fan2024utopiauniversallytrainableoptimal} propose different calibration procedures that are harder to implement in practice, with theoretical coverage guarantees that depend on hyperparameters (see Appendix A.3), (b) we rely on a dual formulation which can handle several hundreds of samples, (c) we propose a goal-oriented tuning strategy for \(\theta^f\), and (d) our proposal is more general and we give better understanding of why this targets adaptivity\footnote{This lack of understanding is for example pointed out by \citet{liang2024conformalpredictionefficiencyorientedmodel}.}. In particular, adaptivity can actually be controlled through the complexity of the scaling function \(f\), in three different ways:
\begin{enumerate}
    \item The sparsity in the linear combination. This can be controlled by the nuclear norm \(\lVert\mathcal{A}\rVert_{\star}\), which acts as a lasso-type penalty (see \citet{recht2010guaranteed}),
    \item The $\ell_2$ norm of the coefficients in the linear combination, which is equal to the Frobenius norm \(\lVert\mathcal{A}\rVert_{F}\) and is similar to a ridge penalty,
    \item The RKHS \(\mathcal{H}^f\) which impacts the functions \(u_{l}(\cdot)\), and notably the lengthscales \(\theta^f\).
\end{enumerate}
To further illustrate point 3., if we take \(k^f(x,x')=\langle x,x'\rangle\) (RHKS of linear functions), the rescaling function will be a second-order polynomial (thus not complex), while with a Gaussian kernel if \(\theta^f\rightarrow +\infty\) the prediction intervals will be constant (not adaptive enough) and if \(\theta^f\) is small they will be very wiggly (too adaptive). We thus see that in between, we can target better adaptivity: we propose in \Cref{subsec:hyper-parameters tuning} a new goal-oriented criterion related to local coverage to tune the lengthscales \(\theta^f\). Note also that \(1-\alpha\) instead of \(100\%\) coverage can be considered in the constraints, but this unfortunately leads to a non-convex optimization problem \citep{braun2025minimum}.

Before discussing in detail the choice of our problem hyperparameters, we first derive a representer theorem which makes the optimization numerically tractable. 
\begin{theorem}[Representer theorem]
\label{thm:representer theorem}
    Let \((a,b,s,\lambda_1)\in\mathbb{R}_{+}^{4}\) and \(\lambda_2>0\).
    Then \Cref{eq:infdim1} admits a unique solution \((m^{\star},f_{\mathbf{A}^{\star}})\) of the form \(m^{\star}(X) = \sum_{i=1}^{n}\gamma^{\star}_{i}k^{m}(X_i, X)={\bs{\gamma}^{\star}}^{\top}\mathbf{k}^{m}_X\) and \(f_{\mathbf{A}^{\star}}(X) = \mathbf{\Phi}(X)^{\top}\mathbf{A}^{\star}\mathbf{\Phi}(X)\) for some vector \(\bs{\gamma}^{\star}\in \mathbb{R}^{n}\) and matrix \(\mathbf{A}^{\star}\in \mathbb{R}^{n\times n},\mathbf{A}^{\star}\succeq0\).
\end{theorem}
The detailed proof, based on \citet{marteauferey2020nonparametricmodelsnonnegativefunctions} and \citet{muzellec2022learningpsdvaluedfunctionsusing}, can be found in Appendix A.1.
This representer theorem leads to the following tractable semi-definite programming (SDP) problem:
\begin{align}
\label{equation:semi-definite problem}
    \inf_{\bs{\gamma}\in \mathbb{R}^{n},\; \mathbf{A}\in \mathbb{S}_{+}^{n}} \quad&\frac{a}{n}\sum_{i=1}^{n} \left(Y_{i}-\bs{\gamma}^{\top}\mathbf{k}^{m}_{X_i}\right)^{2} + \frac{b}{n}\sum_{i=1}^{n} \tilde{f}_{\mathbf{A}}(X_{i}) + \lambda_{1}\lVert \mathbf{A} \rVert_{\star} + \lambda_{2}\lVert \mathbf{A}\rVert_{F}^{2} \nonumber \\
    \text{s.t.} \quad& \tilde{f}_{\mathbf{A}}(X_{i}) \geq \left(Y_{i}-\bs{\gamma}^{\top}\mathbf{k}^{m}_{X_i}\right)^{2}, \;i \in \left[n\right],  \\
    \quad& \bs{\gamma}^{\top}\mathbf{K}^{m}\bs{\gamma} \leq s \nonumber.
\end{align}  
In practice, such SDP problem can be solved using off-the-shelf solvers like \texttt{SCS} \citep{odonoghue2016conicoptimizationoperatorsplittinghomogeneousselfdualembedding}, as was advocated by \citet{liang2022universalpredictionbandssemidefiniteprogramming} and \citet{fan2024utopiauniversallytrainableoptimal}. However, our numerical experiments show that this strategy does not scale past a few hundreds pre-training samples: this is thus a severe practical limitation which, in our opinion, heavily weakens the kernel SoS point of view. To circumvent this major issue, we rely instead on a dual formulation of \Cref{equation:semi-definite problem}.

\paragraph{Dual formulation.} First note that this formulation is possible only when \(\lambda_2>0\) (thus excluding \citet{liang2022universalpredictionbandssemidefiniteprogramming} framework) and that it consists of an optimization problem over  \((n+1)\) variables rather than \((n+n\times n)\) variables.
\begin{proposition}[Dual formulation]
\label{prop:dual formulation}
    Let \((a,b,s,\lambda_1)\in\mathbb{R}_{+}^{4},\, \lambda_2>0\) and \(\Delta:=\mathbb{R}_+^{n+1}\). \Cref{equation:semi-definite problem} admits a dual formulation of the form
    \begin{align}
    \label{eq:dual function}
        \sup_{(\bs{\Gamma}, \theta) \in \Delta}\mathbf{r}(\bs{\Gamma}, \theta)^{\top}\mathrm{Diag}(\bs{\Gamma}_{\mathbf{a}})\mathbf{r}(\bs{\Gamma}, \theta) +
        \theta (\bs{\gamma}(\bs{\Gamma}, \theta)^{\top} \mathbf{K}^m \bs{\gamma}(\bs{\Gamma}, \theta) - s) -
        \Omega^{\star}(\mathbf{V}\mathrm{Diag}(\bs{\Gamma}_{-\mathbf{b}})\mathbf{V}^{\top})
    \end{align}
    where \(\mathbf{r}(\bs{\Gamma}, \theta) = \mathbf{Y} - \mathbf{K}^m \bs{\gamma}(\bs{\Gamma}, \theta)\), \(\bs{\gamma}(\bs{\Gamma}, \theta) = \mathbf{C}\left(\bs{\Gamma}, \theta\right)^{-1} \mathrm{Diag}(\bs{\Gamma}_{a})\mathbf{Y}\), \(\mathbf{C}\left(\bs{\Gamma}, \theta\right) = \mathrm{Diag}\left(\bs{\Gamma}_a\right)\mathbf{K}^{m}+\theta \mathbf{I}_{n}\), \(\Omega^{\star}(\mathbf{B}) = \frac{1}{4\lambda_{2}}\lVert \left[\mathbf{B}-\lambda_{1}\mathbf{I}_{n}\right]_{+}\rVert_{F}^{2}\) and \(\forall x \in \mathbb{R}, \;\mathrm{Diag}\left(\bs{\Gamma}_x\right) := \mathrm{Diag}\left(\bs{\Gamma}\right) + \frac{x}{n}\mathbf{I}_{n}\).
    Moreover, if \((\widehat{\bs{\Gamma}}, \widehat{\theta})\) is solution of \Cref{eq:dual function}, a solution of \Cref{eq:infdim1} can be retrieved as 
    \begin{align*}
        \widehat{\bs{\gamma}} = \left(\mathrm{Diag}(\widehat{\bs{\Gamma}}_\mathbf{a})\mathbf{K}^{m}+\widehat{\theta}\mathbf{I}_n\right)^{-1}\mathrm{Diag}(\widehat{\bs{\Gamma}}_\mathbf{a})\mathbf{Y} \quad \text{and} \quad
        \widehat{\mathbf{A}} = \frac{1}{2\lambda_{2}}\left[\mathbf{V}\mathrm{Diag}(\widehat{\bs{\Gamma}}_{-\mathbf{b}})\mathbf{V}^{\top}-\lambda_{1}\mathbf{I}_{n}\right]_{+}
    \end{align*}
    where \(\left[\mathbf{A}\right]_{+}\) denotes the positive part of \(\mathbf{A}\)\footnote{For a PSD matrix \(\mathbf{A}\) with eigendecomposition \(\mathbf{A} = \mathbf{U} \mathbf{D} \mathbf{U}^{\top}\), its positive part is defined as \(\left[\mathbf{A}\right]_{+} = \mathbf{U} \max(0,\mathbf{D}) \mathbf{U}^{\top}\).} .
\end{proposition}
A detailed proof is given in Appendix A.2 as well as the dual analytical gradient. Interestingly, this dual formulation can be efficiently optimized using accelerated gradient-ascent algorithms \citep{ruder2017overviewgradientdescentoptimization, xie2024adanadaptivenesterovmomentum}, see Figure \ref{fig:comparisons_large} for an illustration of the numerical speed-up. 

\paragraph{Final prediction bands.} Solving the dual formulation yields estimates \(\widehat{m}(X)=\widehat{\bs{\gamma}}^{\top}\mathbf{k}^{m}_X\) and \(\widehat{f}_{\mathbf{A}}(X)=\mathbf{\Phi}(X)^{\top}\widehat{\mathbf{A}}\mathbf{\Phi}(X)\), from which we derive the estimated score function \(S(X,Y)= \vert Y-\widehat{m}(X)\vert/\sqrt{\widehat{f}_{\mathbf{A}}(X)}\). Following standard split CP, our prediction interval is given by 
\begin{align}
\label{eq:final_interval}
    \widehat{C}_{N}(X_{N+1}) = \left[\widehat{m}(X_{N+1})-\widehat{q}_{\alpha}\sqrt{\widehat{f}_{\mathbf{A}}(X_{N+1})},\widehat{m}(X_{N+1})+\widehat{q}_{\alpha}\sqrt{\widehat{f}_{\mathbf{A}}(X_{N+1})}\right]
\end{align}
where \(\widehat{q}_{\alpha}\) is the adjusted quantile of the estimated score function on the calibration set. By design (see Appendix A.3), they satisfy the same marginal coverage guarantee as split CP, as formalized below.
\begin{proposition}
    If the samples \(\left(X_1, Y_1\right),\ldots,\left(X_N, Y_N\right),\left(X_{N+1}, Y_{N+1}\right)\) are exchangeable, then the prediction interval (\ref{eq:final_interval}) satisfies the marginal coverage (\ref{eq:marginal_coverage}).
\end{proposition}

\paragraph{Extensions.} Similarly to \citet{liang2022universalpredictionbandssemidefiniteprogramming}, if a point estimate \(\widehat{m}(\cdot)\) is available beforehand from any machine learning model trained on a separate dataset, our dual formulation can be easily modified. More importantly, it is also straightforward to consider non-symmetric intervals by estimating two functions \(\widehat{f}_{\mathbf{A}^{\textrm{low}}}(\cdot)\) and \(\widehat{f}_{\mathbf{A}^{\textrm{up}}}(\cdot)\), at the cost of increasing the dual problem dimension to \((2n+1)\), see Appendix B.4. We also strongly believe that incorporating constraints of the form \(\mu_1(\cdot) \leq \mu_2(\cdot)\) as proposed by \citet{xie2024boostedconformalpredictionintervals} is possible with kernel SoS, by taking inspiration from \citet{aubin2024approximation}. Kernel SoS is also complementary to \citet{hore2024conformalpredictionlocalweights}, as our learned score can be post-processed with their approach. Finally, since we aim for \(100\%\) coverage, our framework may lack robustness with respect to outliers: however, we can take advantage of our preliminary estimate \(m_{\textrm{GP}}(\cdot)\) (see next section) to filter out samples with large residuals.

\subsection{Hyperparameter tuning}
\label{subsec:hyper-parameters tuning}

First, the hyperparameters \(\theta^m\) and \(s\) related to the regression function \(m(\cdot)\) are fixed with a preliminary estimate \(m_{\textrm{GP}}(\cdot)\) obtained from Gaussian Process (GP) regression with kernel \(k^m\):  \(\theta^m\) is optimized by maximum likelihood and \(s=\lVert m_{\textrm{GP}}\rVert_{\mathcal{H}^{m}}^{2}\). Second, our numerical experiments in Appendix B.1 show that \(\lambda_2\) has a very small impact on the shape of the intervals, as long as \(\lambda_2>0\). Not only does it make the initial problem strongly convex and makes the dual formulation possible, but it also facilitates numerical optimization. In the following, we thus fix it at \(\lambda_2=1\).

Hyperparameters \(a,b,\lambda_1\) and \(\theta^f\) necessitate more attention. We observe numerically that \(a\) has very little influence on mean squared-errors when the noise is symmetric (see Appendix B.1). This phenomenon was commented in \citet{fan2024utopiauniversallytrainableoptimal}, where they argue that the coverage constraints have the additional effect to reduce mean squared-errors, but here we insist that this is true only for symmetric noise. Our experiments confirm this intuition, we simply set \(a=0\) for the symmetric case. Furthermore, they also show that \(b\) and \(\lambda_1\) have highly different influence. This may seem surprising, given how they interact in the dual formulation through the last term in (\ref{eq:dual function}). A potential explanation is that \(b\) controls a data-driven term, unlike \(\lambda_1\), and the latter does not have a strong impact on the optimal solution provided it is positive. As a result, we choose to set \(\lambda_1=1\). On the contrary, \(b\) has the expected behavior, in the sense that higher values yield narrower intervals, at the cost of increasing the nuclear norm, that is the function complexity. But the last hyperparameter \(\theta^f\) also controls such complexity: we thus expect a compensation between these two hyperparameters through an interaction. Numerical experiments in Section \ref{sec:experiments} actually confirm this phenomenon and show that as long as \(b\) is sufficiently large (\textit{e.g.} \(10\) or \(100\), we recommend to test these two values), it is possible to tune only \(\theta^f\) to reach the same level of adaptivity. \(\theta^f\) thus requires specific consideration, since it has the largest impact when \(b\) is fixed. This is the one we focus on below.

\paragraph{Goal-oriented criterion.}
Adaptivity is tightly related to local coverage \(\mathbb{P}(Y_{N+1}\in \widehat{C}(X_{N+1}) \mid X_{N+1}=x )\geq 1-\alpha\), which is impossible to achieve exactly in a distribution-free setting \citep{vovk2012conditionalvalidityinductiveconformal, barber2020limitsdistributionfreeconditionalpredictive}.
We focus ourselves on a weaker version, where we condition on \(X\) being in a small neighborhood \(\omega_{X}\in\mathcal{F}_X\) from the event space  $\mathcal{F}_X$ such that for all \(x\in\mathcal{X}, \mathbb{P}(x\in \omega_{X})\geq \delta\): \(\mathbb{P}(Y_{N+1}\in\widehat{C}(X_{N+1}) \lvert X_{N+1}\in \omega_{X})\geq 1 - \alpha\).
Recently, \citet{deutschmann2023adaptiveconformalregressionjackknife} showed that such coverage with split CP can be controlled with the mutual information (MI) between the inputs and the score function, namely
\begin{align}
    \label{eq:deutschmann}
    \mathbb{P}(Y\in\widehat{C}_{\mathcal{D}_N}(X) \lvert \mathcal{D}_N, X\in \omega_{X})\geq 1 - \alpha - \frac{1}{\delta}\sqrt{1-\exp(-\textrm{MI}(X,S_{\mathcal{D}_n}(X,Y)))}.
\end{align}
Note that contrary to \citet{deutschmann2023adaptiveconformalregressionjackknife}, we explicitly write the coverage conditionally on the training set $\mathcal{D}_N$. Inherently this bound is uninformative for low-probability
sets, but interestingly we can counterbalance this effect by choosing a score function which is as independent as possible of the inputs in order to get \(\textrm{MI}(X,S(X,Y))\) close to \(0\), see \citet{deutschmann2023adaptiveconformalregressionjackknife}. However, this implies computing MI for a random vector in dimension \(d\): from a practical perspective, MI suffers from the curse of dimensionality and rapidly becomes numerically unstable. Instead, for a re-scaled score function, we show that a similar bound holds with MI between one-dimensional variables, which is more robust. Furthermore, using recent inequalities between the total variation distance and the maximum mean discrepancy \citep{wang2023seminonparametricestimationdistributiondivergence}, we also extend our result to replace MI with HSIC \citep{gretton2005measuring}, for which we observe in practice much improved numerical stability.

\begin{proposition}
\label{prop:hsic}
    Let \(\widehat{C}_{\mathcal{D}_N}\) be the prediction intervals built from a score function \(S(X,Y)=\vert Y-m(X)\vert/\sqrt{f(X)}\) through split CP with \(\mathcal{D}_N=\mathcal{D}_n\cup \mathcal{D}_m\). Then for any \(\omega_X\) in \(\mathcal{F}_X\) such that \(\mathbb{P}(X\in \omega_{X})\geq \delta\), denoting \(p_{\mathcal{D}_N}=\mathbb{P}(Y_{N+1}\in\widehat{C}_{\mathcal{D}_N}(X_{N+1}) \lvert \mathcal{D}_N,X_{N+1}\in \omega_{X})\) we have:
    \begin{align}
    \label{eq:midim1}
        p_{\mathcal{D}_N} \geq 1 - \alpha - \frac{1}{\delta}\sqrt{1 - \alpha_1 \exp(\mathrm{MI}(r_{\mathcal{D}_n}(X_{N+1},Y_{N+1}),\widehat{f}_{\mathcal{D}_n}(X_{N+1})))}
    \end{align}
    where \(\alpha_1\) does not depend on \(f(\cdot)\) and \(r_{\mathcal{D}_n}(X,Y) = \vert Y-\widehat{m}_{\mathcal{D}_n}(X)\vert\). In addition, we have  
    \begin{align}
    \label{eq:hsicdim1}
        p_{\mathcal{D}_N} \geq 1 - \alpha - \frac{1}{\delta}\sqrt{1-\frac{\alpha_1}{1-\alpha_2\mathrm{HSIC}(r_{\mathcal{D}_n}(X_{N+1},Y_{N+1}),\widehat{f}_{\mathcal{D}_n}(X_{N+1}))}}
    \end{align}
    where \(\alpha_2\) only depends on the kernel used for HSIC, which must be characteristic.
\end{proposition}
To target local coverage, we can then maximize \(\textrm{HSIC}(r(X,Y),f(X))\): in our kernel SoS procedure, this means that \(\theta^f\) can be tuned efficiently according to this criterion. For HSIC estimation, we need samples \(X_{N+1}\) independent of \(\mathcal{D}_n\) and thus rely on a cross-validation procedure. Finally, to handle cases where the noise is homoscedastic, we implement as a last step a HSIC test of independence: if the p-value is large, we set \(\theta^f\) to an arbitrary large value. See Appendix A.4 for the proof and B.2 for implementation details.

\begin{remark}
    We believe that \Cref{prop:hsic} is also of interest beyond kernel SoS, in the sense that it provides a principled way to tune hyperparameters in all score functions, such as in \citet{hore2024conformalpredictionlocalweights} or \citet{braun2025minimum} for example. However, its true potential would lie in generalizing \Cref{eq:hsicdim1} to other score functions, which is left as future work. It could also be directly used as a loss function in \citet{xie2024boostedconformalpredictionintervals}, instead of estimating local coverage.
\end{remark}

\section{Experiments}
\label{sec:experiments}

\subsection{Small data experiments}

For all experiments, we consider a Matérn $5/2$ kernel for \(k^m\) and \(k^f\). Our proposed kernel SoS algorithm is first compared in the small data regime to standard competitors for split CP: CQR (with random forests, following experiments from \citet{romano2019conformalizedquantileregression,hore2024conformalpredictionlocalweights}) and rescaled scores. For the latter, we only consider two variants of GP for fair comparison, since we also place ourselves in the RKHS setting. We focus on a homoscedastic and heteroscedastic \citep{binois2018practical} GP model, and consider here \(Y=m(X)+\sigma(X)\epsilon\) with:

\begin{center}
\begin{tabular}{c|c|c|c|c}
& \(m(X),\ Z=10X+1\) & \(\sigma(X)\) & X & \(\epsilon\) \\
\hline
\textrm{Case 1} & \(\sin(\frac{2\pi Z}{5} + 0.2\frac{4\pi Z}{5})\mathds{1}_{X\leq 9.6}+(\frac{Z}{10}-1)\mathds{1}_{X> 9.6}\) & \(\sqrt{0.1+2X^2}\) & \(\mathcal{U}[-1,1]\) & \(\mathcal{N}(0,1)\)\\
\textrm{Case 2} & \(X/2\) & \(\vert \sin(X)\vert\) & \(\mathcal{N}(0,1)\) & \(\mathcal{N}(0,1)\)\\
\end{tabular}
\end{center}

\begin{figure}[ht!]
    \centering
    \[\vcenter{\hbox{\includegraphics[scale=0.15]{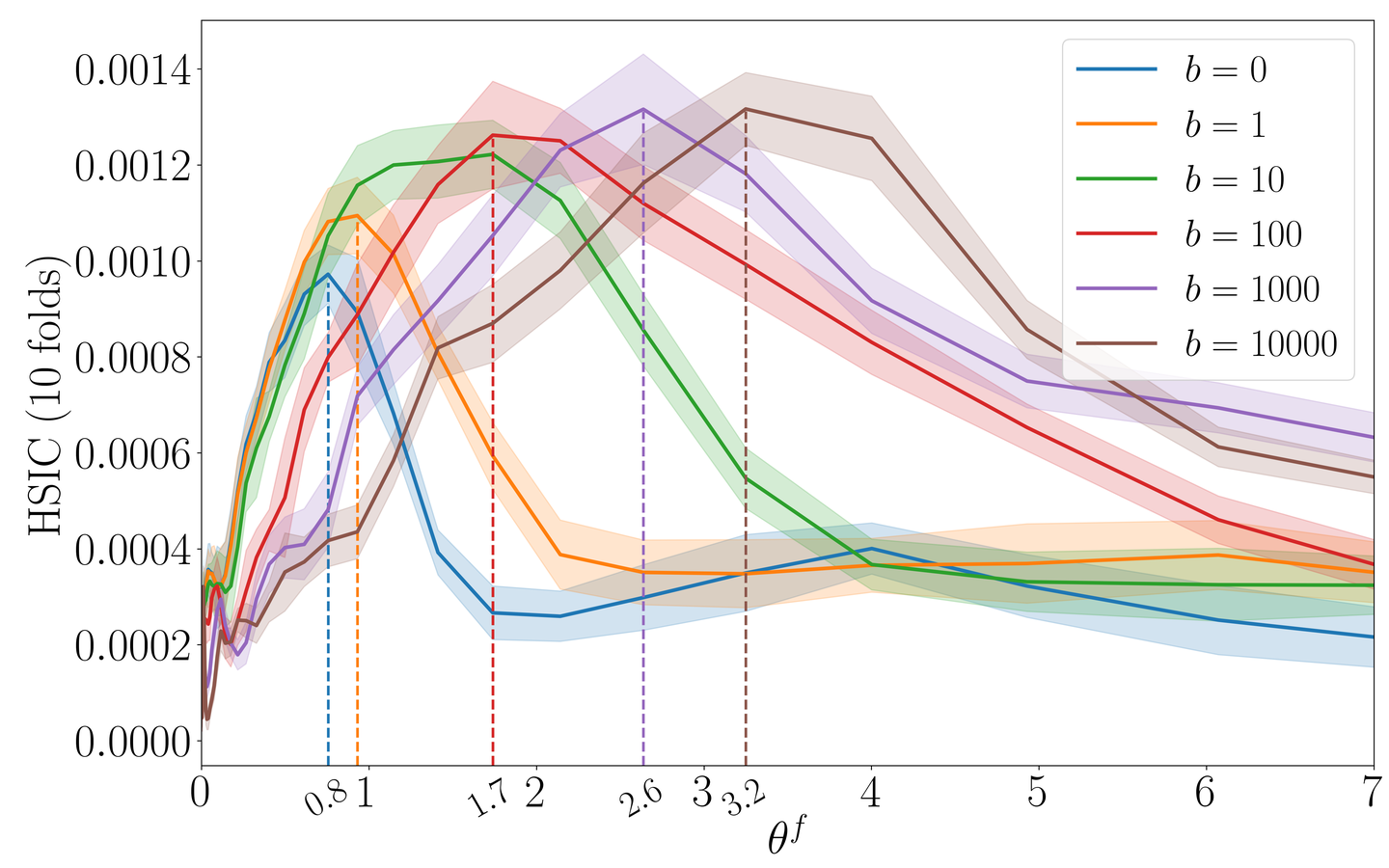}}}
    \vcenter{\hbox{\includegraphics[scale=0.15]{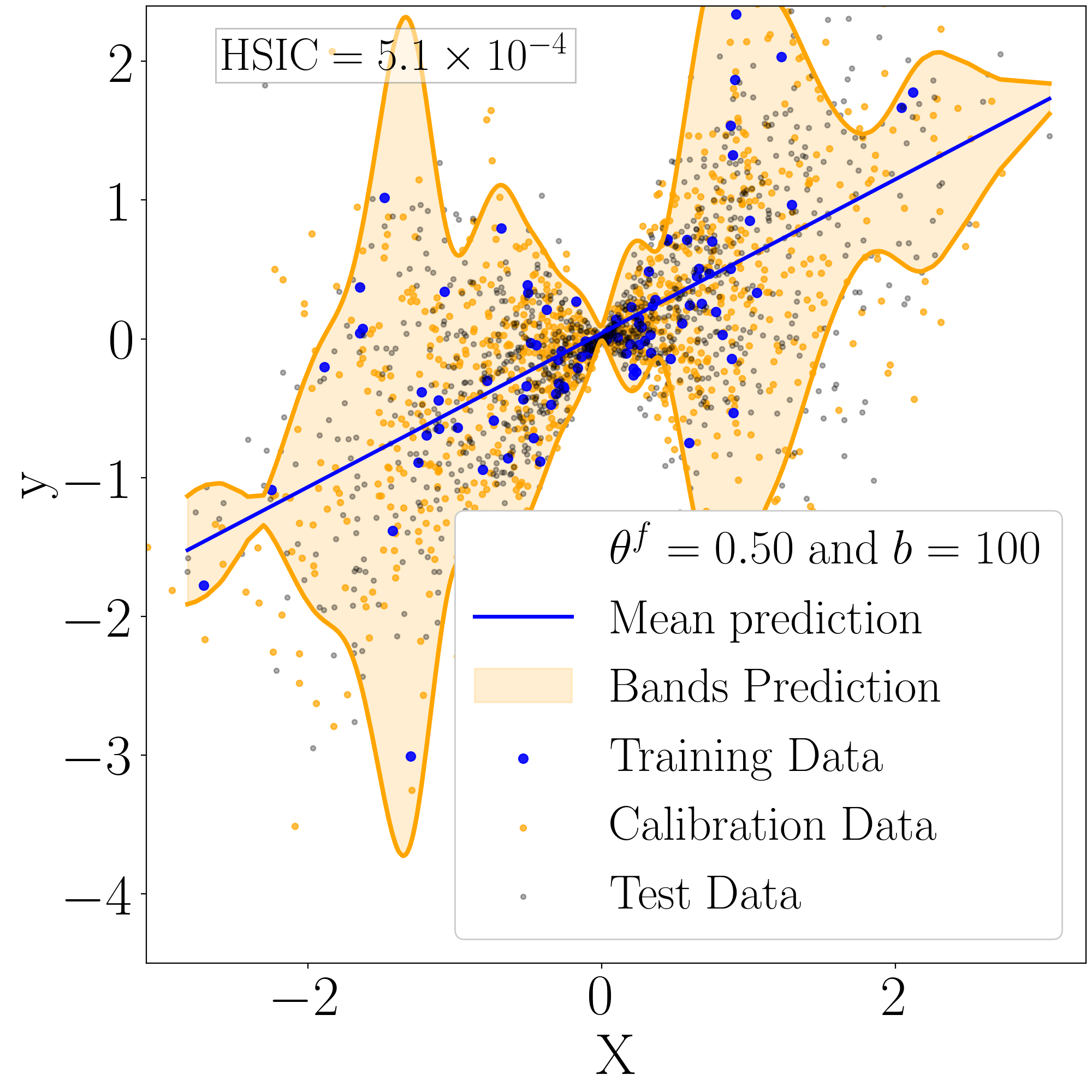}}}
    \vcenter{\hbox{\includegraphics[scale=0.15]{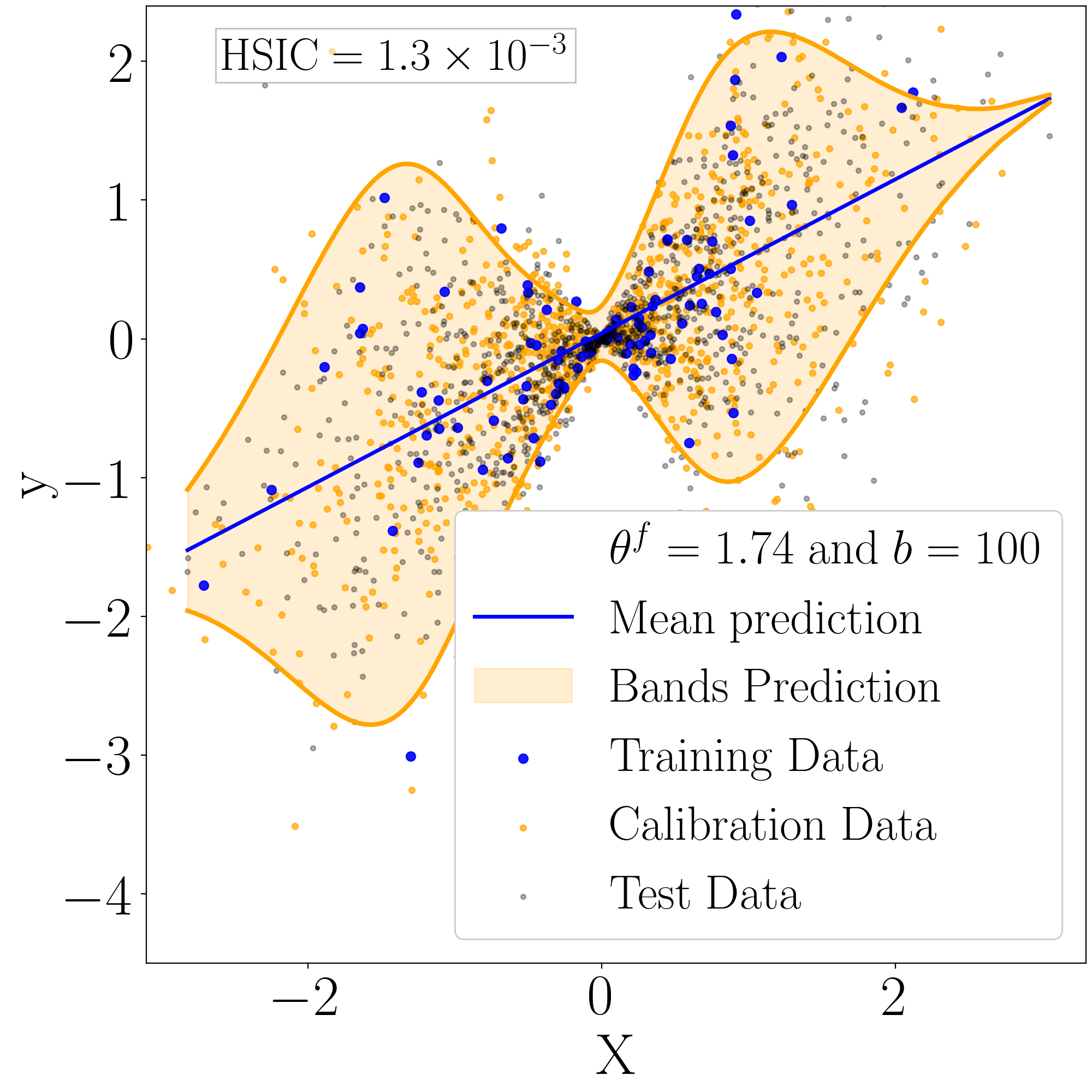}}}\]
    \caption{Test case 2 with \(n=100\). Left: HSIC criterion between \(r(X,Y)\) and \(f(X)\) as a function of \(b\) and \(\theta^f\) (confidence intervals obtained by bootstrap and optimal values of \(\theta^f\) in dashed lines). Middle / Right: optimal prediction bands with too small and optimized lengthscale, respectively.}
    \label{fig:HSIC_versus_b_thetaf}
\end{figure}

We begin by illustrating the interaction between \(b\) and \(\theta^f\) in Figure \ref{fig:HSIC_versus_b_thetaf} left. For all values of \(b\), we observe a consistent HSIC behaviour: it first increases with \(\theta^f\), thus improving adaptivity, until it reaches a peak and then decreases. Interestingly, we also note that the higher \(b\), the higher the optimal \(\theta^f\): this clearly shows that both hyperparameters have opposite effects on adaptivity. Furthermore, we see that for \(b\geq 10\) we reach a plateau for the optimal HSIC. In practice it is thus sufficient to only optimize \(\theta^f\) as soon as \(b\) is fixed at a large enough value. Figure \ref{fig:HSIC_versus_b_thetaf} also shows that a small value for \(\theta^f\) leads to overly adaptive bands, while the HSIC-optimized \(\theta^f\) produces smooth and adaptive ones.

\begin{figure}[ht!]
\centering
\begin{tabular}{cc}
    \begin{minipage}{0.5\textwidth} 
    \includegraphics[width=0.7\textwidth,height=0.88in]{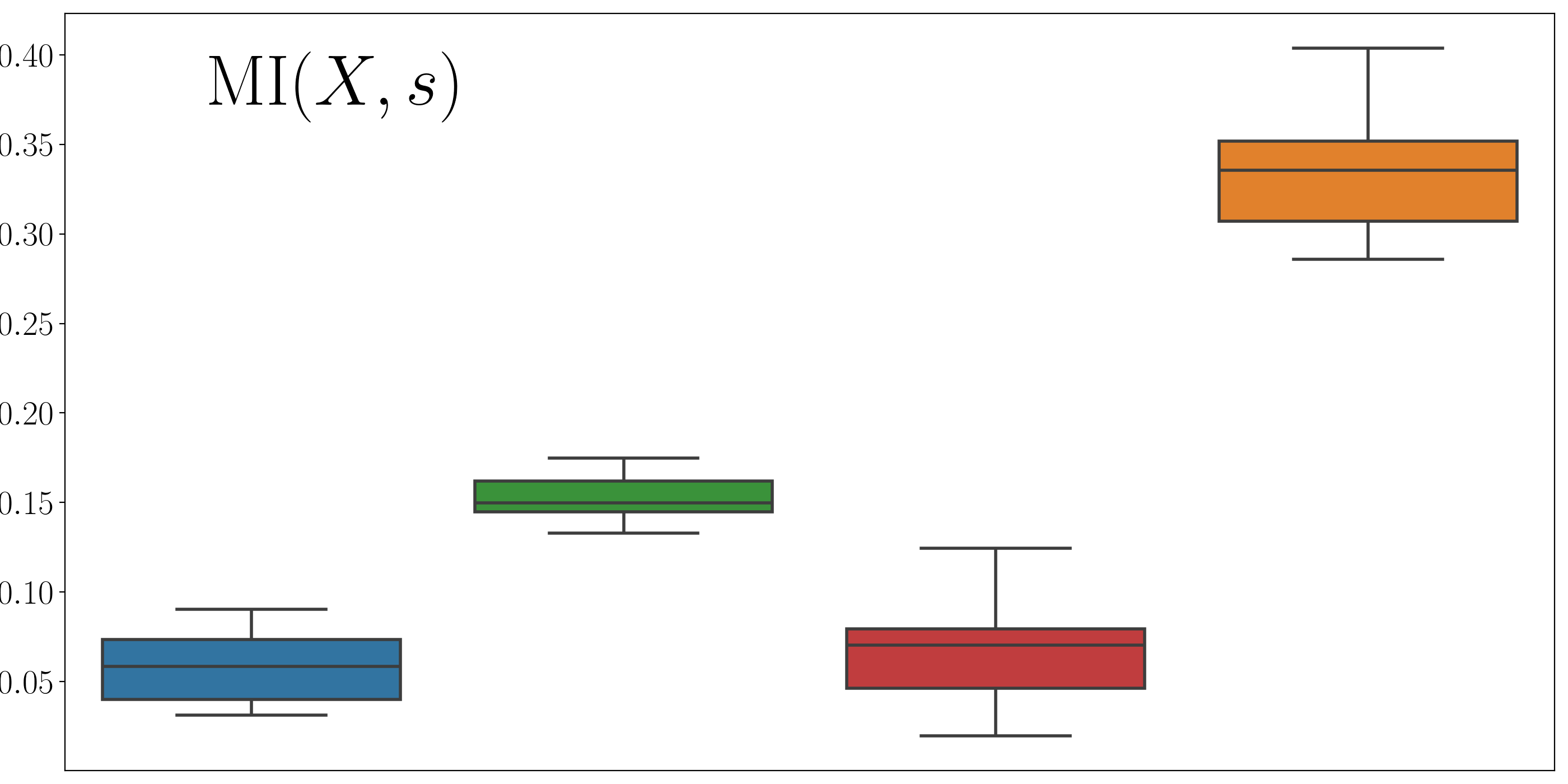} \\ 
    \includegraphics[width=0.7\textwidth,height=0.88in]{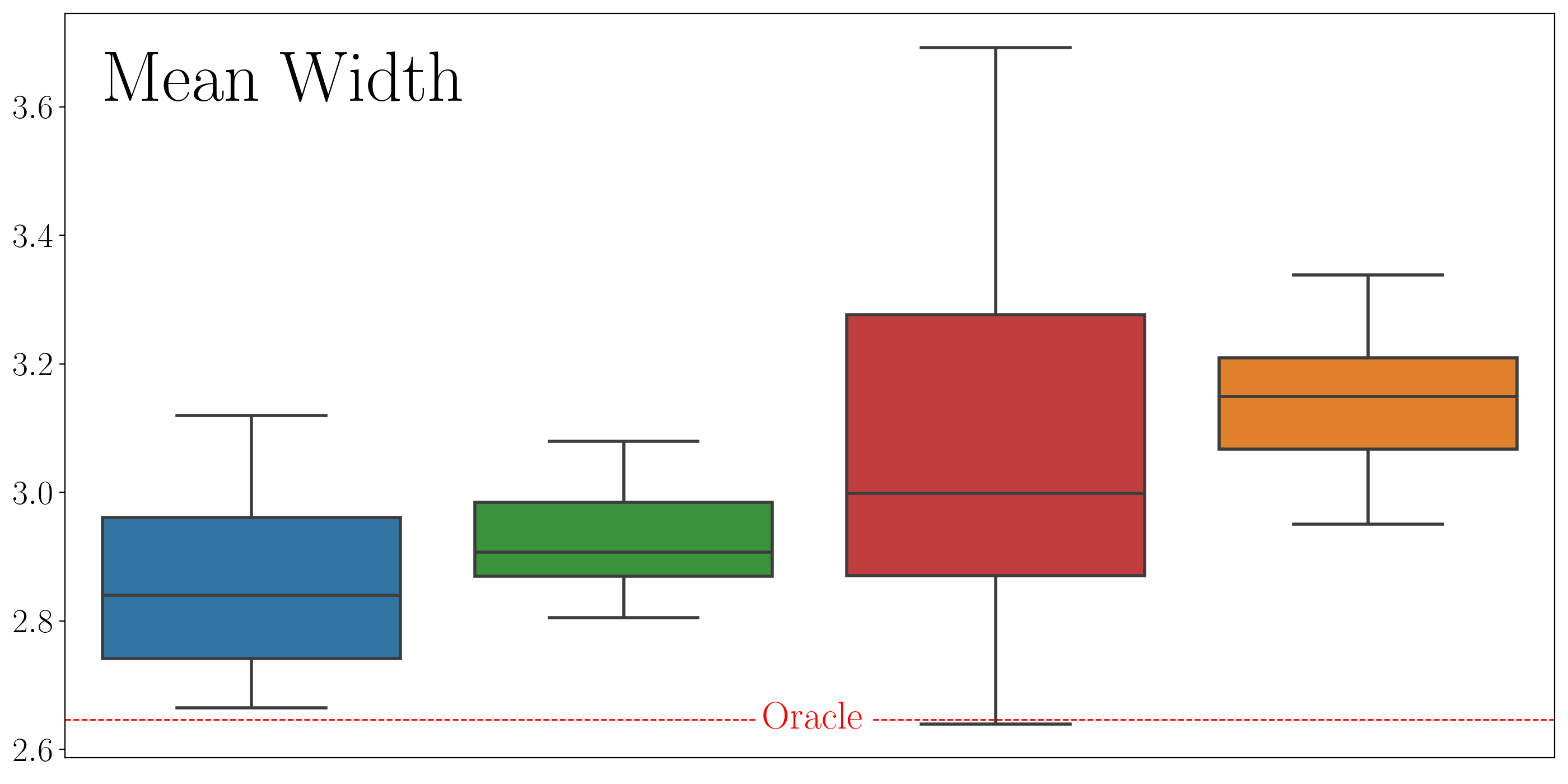}\end{minipage} &  
    \begin{minipage}{0.5\textwidth} 
    \hspace{-23mm}
    \includegraphics[width=1.2\textwidth,height=1.77in]
    {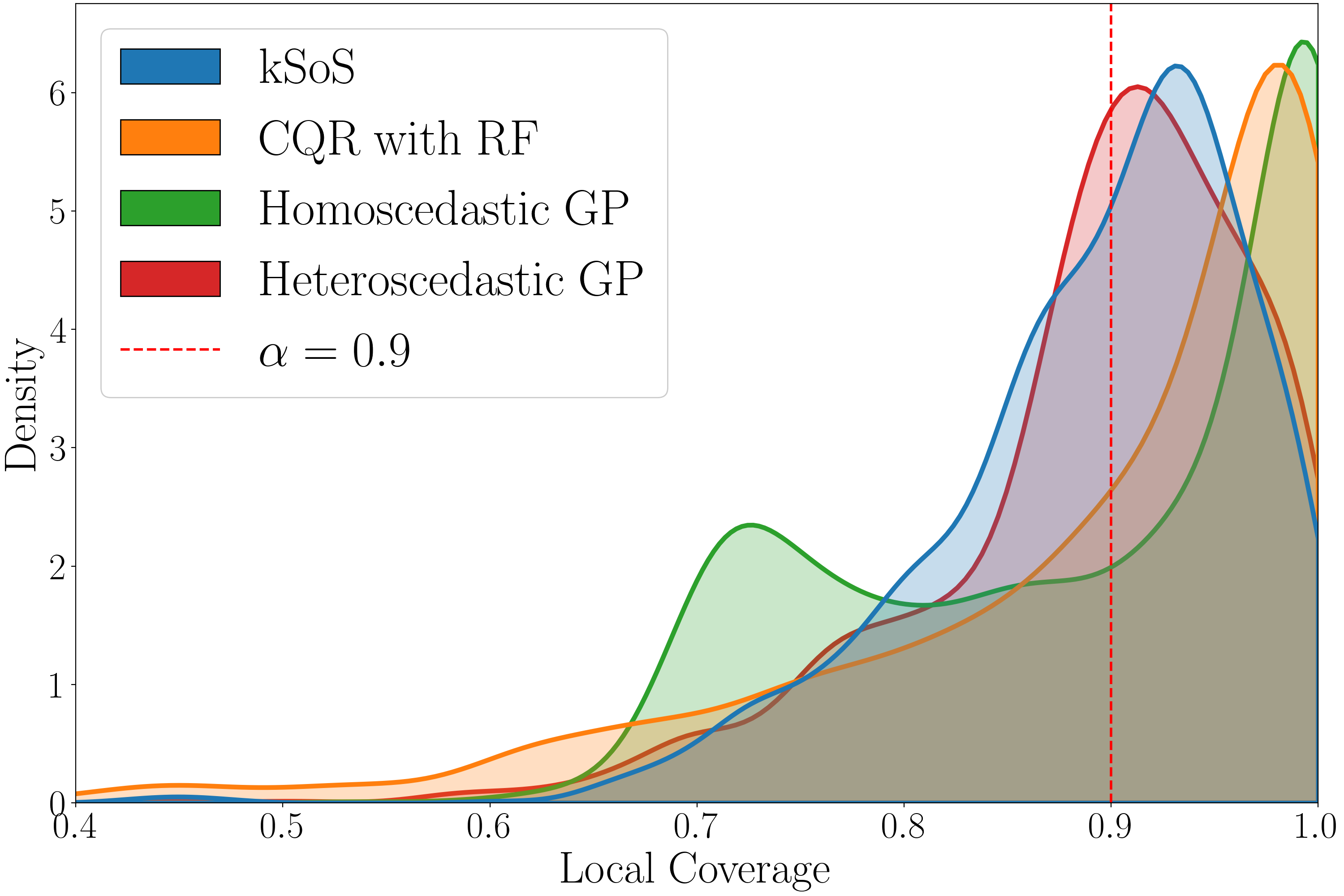}
    \end{minipage} 
    \end{tabular}
\caption{Test case 1 with \(n=100\). Adaptivity metrics and density of local coverage.}
\label{fig:comparisons}
\end{figure}

In the following we now fix \(b=10\) and compare kernel SoS with HSIC-optimized \(\theta^f\) to CQR and rescaled GPs  on test case 1 with $20$ replications. In Figure \ref{fig:comparisons}, we investigate several metrics related to adaptivity: mean width, MI and local coverage, see also Appendix B.4 for complementary discussion. We note that CQR and standard GP yield larger intervals and local coverage with great variability around the target. Heteroscedastic GP produces intervals similar to ours, but with higher mean width. In contrast, kernel SoS gives prediction intervals with both small width and satisfying local coverage properties. Additional experiments confirming this behavior are to be found in Appendix B.4.

\subsection{Real-world and large scale experiments}

\begin{figure}[ht]
\centering
\includegraphics[width=0.45\textwidth]{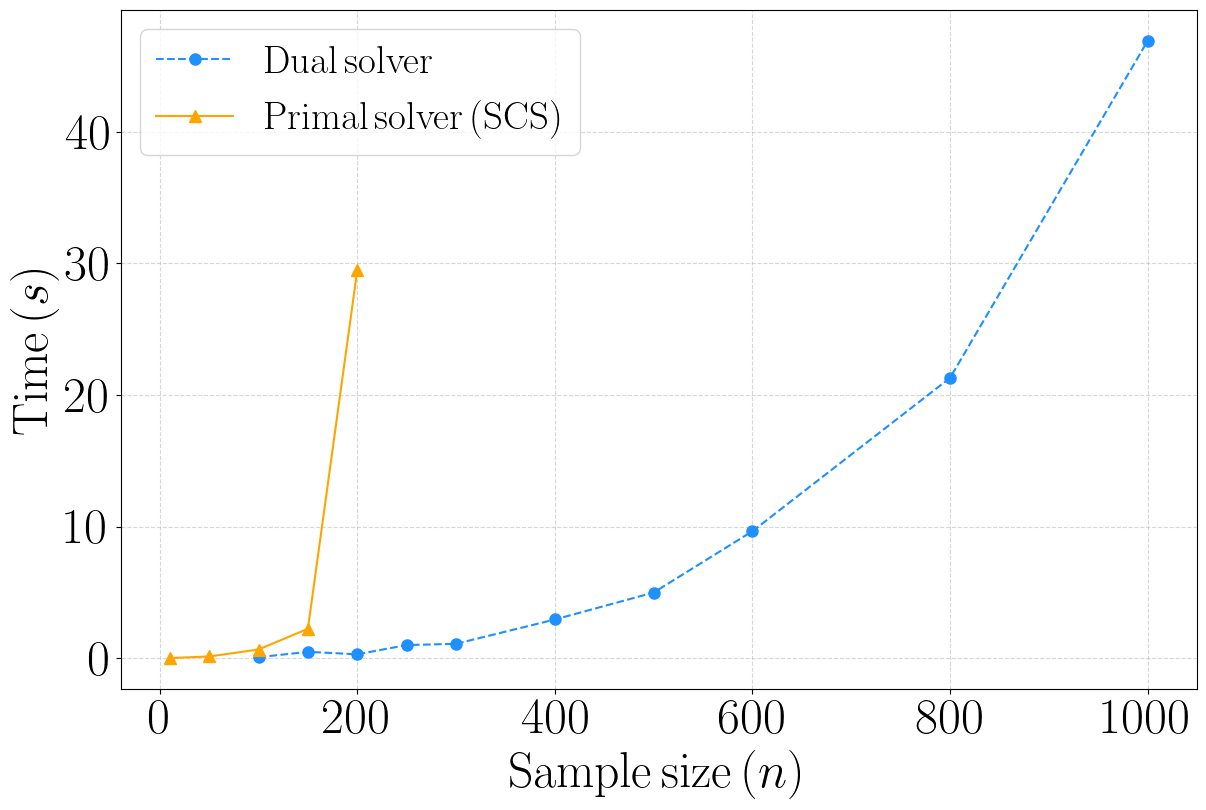}
\includegraphics[width=0.45\textwidth]{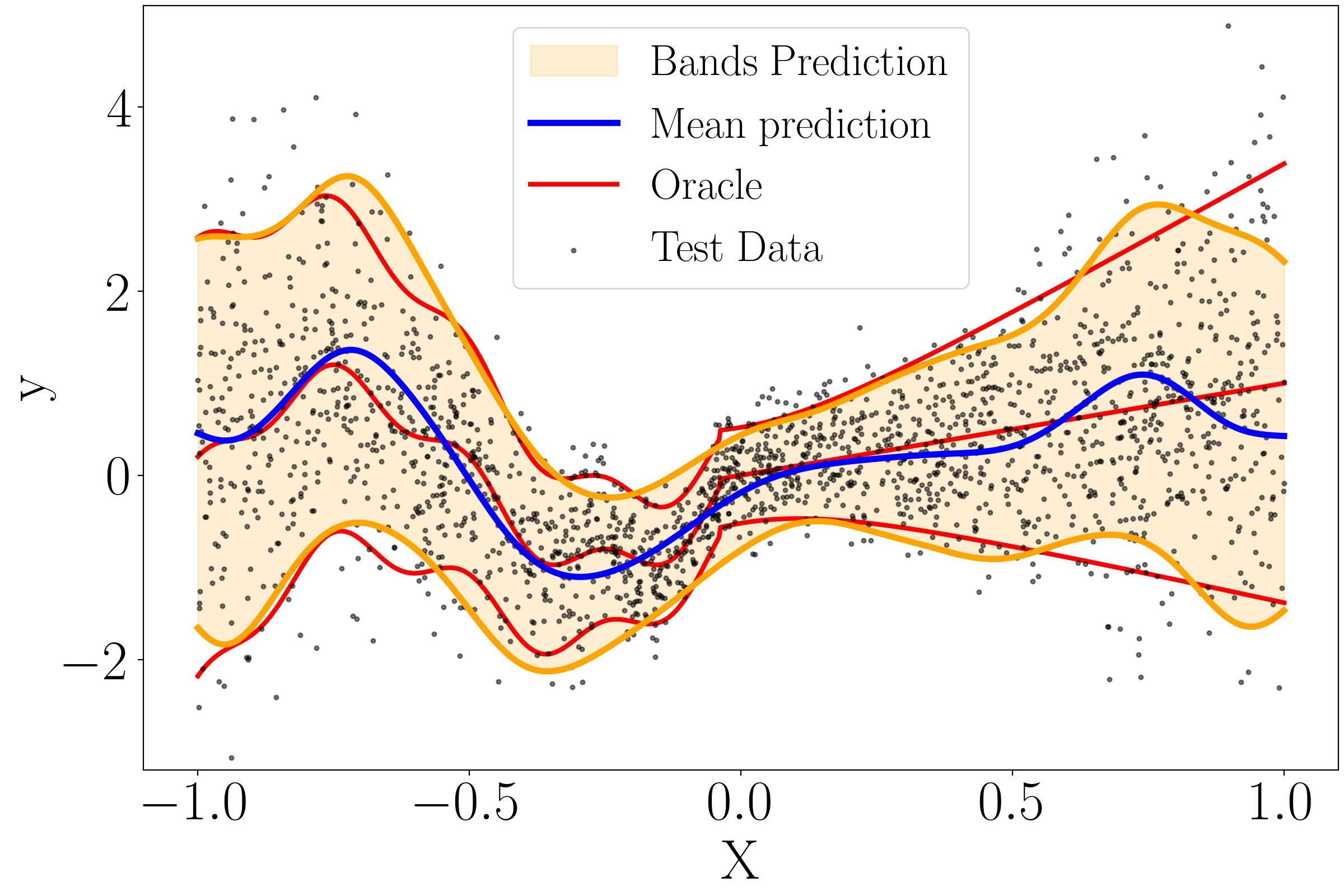}
\caption{Test case 1. Left: time for SDP and dual formulation as a function of \(n\) ($a=b=0,\, \theta^{f}=0.3,\, \textrm{max iter}=10^4$). Right: optimal solution of dual formulation for \(n=2000\).}
\label{fig:comparisons_large}
\end{figure}

Finally, we demonstrate that our dual formulation for solving the kernel SoS problem can scale to several hundreds or thousands of training points, unlike previous work that relies on SDP solvers. Figure \ref{fig:comparisons_large} left, shows that a SDP solver can only handle up to \(n=200\) training samples, while our dual solver scales easily to \(n=1000\) (see Appendix B.3). When we optimize \(\theta^f\) with HSIC, we retrieve the optimal solution in Figure \ref{fig:comparisons_large} right.

At last, we consider six real-world datasets commonly used for regression, which are detailed in Appendix B.4. Table \ref{tab:mean_width_realworld} reports the mean width of prediction intervals on a test set for all methods. We consider two variants of kSoS: one where \(\theta^f\) is chosen to minimize the mean width, and the other where we optimize it with HSIC. Note that for Bio, which exhibits strong asymmetric noise, we use our asymmetric kSoS variant and do not use the HSIC criterion to tune \(\theta^f\), since at the moment it only applies to symmetric intervals. Such asymmetric noise heavily favors CQR, which outperforms all methods on this example. For all other cases except for Concrete which we comment below, kernel SoS targeting mean width achieves better adaptivity when measured by mean width. No other competitor, although they generally have strong performance, achieves this robustness. For completeness, we also provide the marginal coverage for all datasets and methods in Appendix B.4.

\begin{table}[h]
\centering
\resizebox{\textwidth}{!}{%
\begin{tabular}{l|ccccc}
\toprule
Dataset & CQR & Het GP & Hom GP & kSoS & kSoS\\
 &  &  &  & Best mean width & Opt. HSIC \\
\midrule
Concrete   & $0.586 \pm 0.032$ & $\mathbf{0.508} \pm 0.052$ & $0.543 \pm 0.044$ & $0.556 \pm 0.044$ & $0.568 \pm 0.06$\\
Bike       & $1.114 \pm 0.062$ & $1.000 \pm 0.079$ & $\mathbf{0.809} \pm 0.024$ & $\mathbf{0.803} \pm 0.031$ & $\mathbf{0.803} \pm 0032$\\
Bio        & $\mathbf{1.879} \pm 0.046$ & $2.21 \pm 0.100$ & $2.194 \pm 0.119$ & $2.03 \pm 0.07$ & $-$\\
Diabetes   & $188.62 \pm 9.33$ & $191.24 \pm 11.95$ & $190.58 \pm 11.19$ & $\mathbf{185.83} \pm 14.47$ & $187.6 \pm 16.18$\\
MPG        & $9.89 \pm 0.82$ & $9.70 \pm 1.06$ & $9.71 \pm 0.73$ & $\mathbf{9.15} \pm 0.8$ & $9.36 \pm 0.82$\\
Housing    & $1.816 \pm 0.045$ & $1.585 \pm 0.099$ & $\mathbf{1.453} \pm 0.068$ & $\mathbf{1.468} \pm 0.094$ & $1.586 \pm 0.104$\\
\bottomrule
\end{tabular}
}
\caption{Mean width of prediction intervals for six real-world datasets (mean$\pm$sd on 10 repetitions).}
\label{tab:mean_width_realworld}
\end{table}

\begin{figure}[ht]
\centering
\includegraphics[width=0.9\textwidth]{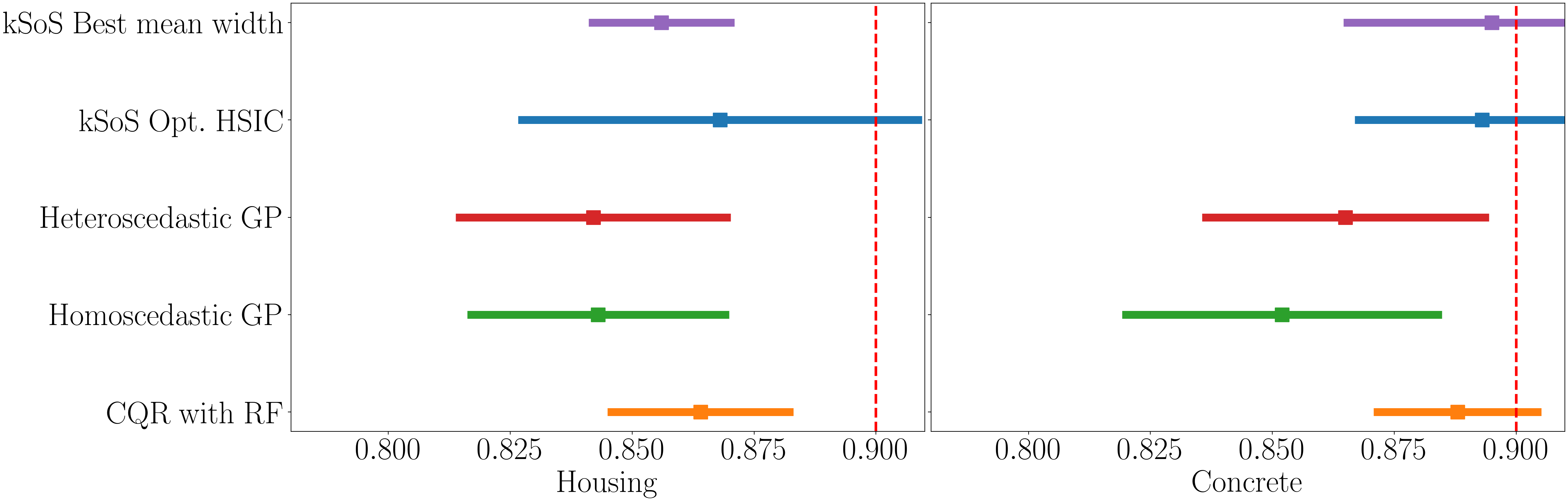}
\caption{Housing (left, with $b=100$) and Concrete (right, with $b=10$) datasets. Mean and standard deviation of worst-set coverage on 10 regions with 100 test points each, 10 repetitions.}
\label{fig:minRC_concrete_housing}
\end{figure}

At first sight, the kSoS variant based on HSIC produces larger bands and appears to be less adaptive if only mean width is considered. However, a more meaningful criterion to evaluate adaptivity is local coverage. Since it is unfortunately out of reach for real datasets, we rely instead on the worst-set coverage \(\min_{l=1,\ldots,L} \mathbb{P}(Y_{N+1}\in\widehat{C}_{\mathcal{D}_N}(X_{N+1}) \lvert X_{N+1}\in \mathcal{R}_l)\) where \(\{\mathcal{R}_l\}_{l=1,\ldots,L}\) is a partition of the input space, see \citet{thurin2025optimal}. We observe in Figure \ref{fig:minRC_concrete_housing} that kSoS with HSIC stays closer to the target $\alpha=0.9$ than all other methods. For Concrete, heteroscedastic GP has the smaller mean width, but at the cost of producing intervals with lower worst-set coverage. Similarly, on Housing both homoscedastic GP and kSoS focusing on mean width exhibit the lowest mean width, but are further from the target. This illustrates once again that our HSIC criterion focuses on local adaptivity. 

\section{Limitations}

Although our experiments show that kernel SoS is a competitor worth trying to build adaptive bands in a moderate sample size regime, we can still identify limitations that would necessitate extensions:
\begin{itemize}
    \item Kernel SoS performs well for dimensions up to 10 or 15, a limitation shared by all kernel methods without specific strategies. For high-dimensional objects with structure (time series, probability distributions, molecules, strings, graphs,….), it is however straightforward to plug specifically designed kernels in our framework. On the other hand, for high-dimensional tabular data, a common strategy is to rely on additive kernels, which are left as future work.
    \item Kernel SoS can scale up to 2000 samples thanks to the dual formulation, but handling much more samples than that would be highly time consuming. This is inherently a limitation that comes from kernels rather than from the dual formulation. Indeed, each objective function and gradient computation involves the usual kernel regression formulas, which scale as $O(n^3)$. A usual workaround is to use Nystrom-type approximations or Random Fourier Features, which could be easily integrated in our problem.
    \item Due to the $100\%$ coverage constraint, kernel SoS is less robust to outliers than CQR. In such cases, a pre-processing step using another model (e.g. the initial GP model we use in our procedure or an heteroscedastic GP model) is recommended.
\end{itemize}

\section{Conclusion}

We introduce a generalized kernel sum-of-squares framework for building scalable and adaptive prediction bands for conformal prediction. Scalability is achieved through a new representer theorem together with a dual formulation which can be solved efficiently with accelerated gradient algorithms. Unlike previous work, this makes the kernel SoS paradigm for CP able to scale up to thousands of training points, as we illustrate in our experiments. On the other hand, adaptivity and local coverage are targeted by optimizing hyperparameters with a new HSIC-based criterion, which is numerically robust and has excellent practical performance. Since such a criterion appears promising, as a perspective, we plan to investigate its extension to more general score functions.


\newpage

\bibliography{bibliography}
\newpage
\section*{NeurIPS Paper Checklist}

\begin{enumerate}

\item {\bf Claims}
    \item[] Question: Do the main claims made in the abstract and introduction accurately reflect the paper's contributions and scope?
    \item[] Answer: \answerYes{} 
    \item[] Justification: We rely on numerical experiments in the main paper to support our claims, and also provide detailed proofs and additional experiments in the Appendix.

\item {\bf Limitations}
    \item[] Question: Does the paper discuss the limitations of the work performed by the authors?
    \item[] Answer: \answerYes{} 
    \item[] Justification: We comment on the limitations in several remarks and in the final section that guide towards future work. We also provide a discussion on computational efficiency and how it scales with sample size in the main paper and the Appendix.

\item {\bf Theory assumptions and proofs}
    \item[] Question: For each theoretical result, does the paper provide the full set of assumptions and a complete (and correct) proof?
    \item[] Answer: \answerYes{} 
    \item[] Justification: Complete proofs and assumptions are available in the Appendix.

    \item {\bf Experimental result reproducibility}
    \item[] Question: Does the paper fully disclose all the information needed to reproduce the main experimental results of the paper to the extent that it affects the main claims and/or conclusions of the paper (regardless of whether the code and data are provided or not)?
    \item[] Answer: \answerYes{} 
    \item[] Justification: We always explicitly mention the hyperparameter values used in all experiments, and also provide reproducible code in the supplemental material.

\item {\bf Open access to data and code}
    \item[] Question: Does the paper provide open access to the data and code, with sufficient instructions to faithfully reproduce the main experimental results, as described in supplemental material?
    \item[] Answer: \answerYes{} 
    \item[] Justification: The accompanying code can be used to reproduce all our results.

\item {\bf Experimental setting/details}
    \item[] Question: Does the paper specify all the training and test details (e.g., data splits, hyperparameters, how they were chosen, type of optimizer, etc.) necessary to understand the results?
    \item[] Answer: \answerYes{} 
    \item[] Justification: We provide in the Appendix an entire section dedicated to all the experimental setting.

\item {\bf Experiment statistical significance}
    \item[] Question: Does the paper report error bars suitably and correctly defined or other appropriate information about the statistical significance of the experiments?
    \item[] Answer: \answerYes{} 
    \item[] Justification: When applicable, we always add error bars or boxplots on our experiments. We also describe the factors of variability.

\item {\bf Experiments compute resources}
    \item[] Question: For each experiment, does the paper provide sufficient information on the computer resources (type of compute workers, memory, time of execution) needed to reproduce the experiments?
    \item[] Answer: \answerYes{} 
    \item[] Justification: We detail in the Appendix the computer resources we used for all our experiments.
    
\item {\bf Code of ethics}
    \item[] Question: Does the research conducted in the paper conform, in every respect, with the NeurIPS Code of Ethics \url{https://neurips.cc/public/EthicsGuidelines}?
    \item[] Answer: \answerYes{} 
    \item[] Justification: Our work has no potential harmful consequence, and as far as data are concerned, we only used synthetic and publicly available datasets.

\item {\bf Broader impacts}
    \item[] Question: Does the paper discuss both potential positive societal impacts and negative societal impacts of the work performed?
    \item[] Answer: \answerNA{} 
    \item[] Justification: NA.
    
\item {\bf Safeguards}
    \item[] Question: Does the paper describe safeguards that have been put in place for responsible release of data or models that have a high risk for misuse (e.g., pretrained language models, image generators, or scraped datasets)?
    \item[] Answer: \answerNA{} 
    \item[] Justification: We only use synthetic datasets.

\item {\bf Licenses for existing assets}
    \item[] Question: Are the creators or original owners of assets (e.g., code, data, models), used in the paper, properly credited and are the license and terms of use explicitly mentioned and properly respected?
    \item[] Answer: \answerNA{} 
    \item[] Justification: We do not rely on existing assets.

\item {\bf New assets}
    \item[] Question: Are new assets introduced in the paper well documented and is the documentation provided alongside the assets?
    \item[] Answer: \answerNA{} 
    \item[] Justification: NA.

\item {\bf Crowdsourcing and research with human subjects}
    \item[] Question: For crowdsourcing experiments and research with human subjects, does the paper include the full text of instructions given to participants and screenshots, if applicable, as well as details about compensation (if any)? 
    \item[] Answer: \answerNA{} 
    \item[] Justification: NA.

\item {\bf Institutional review board (IRB) approvals or equivalent for research with human subjects}
    \item[] Question: Does the paper describe potential risks incurred by study participants, whether such risks were disclosed to the subjects, and whether Institutional Review Board (IRB) approvals (or an equivalent approval/review based on the requirements of your country or institution) were obtained?
    \item[] Answer: \answerNA{} 
    \item[] Justification: NA.

\item {\bf Declaration of LLM usage}
    \item[] Question: Does the paper describe the usage of LLMs if it is an important, original, or non-standard component of the core methods in this research? Note that if the LLM is used only for writing, editing, or formatting purposes and does not impact the core methodology, scientific rigorousness, or originality of the research, declaration is not required.
    \item[] Answer: \answerNA{} 
    \item[] Justification: NA.

\end{enumerate}

\end{document}